\documentclass[12pt]{article}
\usepackage[margin=1in]{geometry}
\usepackage[utf8]{inputenc}
\usepackage[T1]{fontenc}
\usepackage{hyperref}
\usepackage{url}
\usepackage{amsfonts}
\usepackage{nicefrac}
\usepackage{microtype}
\usepackage{graphicx}
\usepackage{float}
\usepackage[justification=centering]{caption}
\usepackage{todonotes}
\usepackage[numbers]{natbib}
\usepackage{booktabs}
\usepackage{multirow}
\usepackage{amsmath}
\usepackage{textcomp}
\usepackage{ragged2e}

\usepackage{framed,multirow}

\usepackage{amssymb}
\usepackage{latexsym}
\usepackage{amsmath}
\usepackage{textcomp}
\usepackage{ragged2e}
\usepackage{todonotes}

\usepackage{url}
\usepackage{xcolor}
\definecolor{newcolor}{rgb}{.8,.349,.1}
\usepackage{graphicx}
\usepackage{booktabs}   % Nicer Tables
\usepackage{float}

\title{Comparing interpretability and explainability for feature selection}

\author{Jack Dunn, Luca Mingardi, Ying Daisy Zhuo\\
        Interpretable AI\\
        Cambridge, MA 02142\\
        \texttt{info@interpretable.ai} \\
}

\date{}

\begin{document}
\maketitle

\begin{abstract}
A common approach for feature selection is to examine the variable importance scores for a machine learning model, as a way to understand which features are the most relevant for making predictions. Given the significance of feature selection, it is crucial for the calculated importance scores to reflect reality. Falsely overestimating the importance of irrelevant features can lead to false discoveries, while underestimating importance of relevant features may lead us to discard important features, resulting in poor model performance. Additionally, black-box models like XGBoost provide state-of-the art predictive performance, but cannot be easily understood by humans, and thus we rely on variable importance scores or methods for explainability like SHAP to offer insight into their behavior.

In this paper, we investigate the performance of variable importance as a feature selection method across various black-box and interpretable machine learning methods. We compare the ability of CART, Optimal Trees, XGBoost and SHAP to correctly identify the relevant subset of variables across a number of experiments. The results show that regardless of whether we use the native variable importance method or SHAP, XGBoost fails to clearly distinguish between relevant and irrelevant features. On the other hand, the interpretable methods are able to correctly and efficiently identify irrelevant features, and thus offer significantly better performance for feature selection.
\end{abstract}

%\linenumbers

%% main text
\section{Introduction}
\label{intro}
In the modern era of the Internet of Things, data is created and collected every day at an ever-increasing rate, leading to datasets with many thousands of characteristics associated with each data point. Examples include detailed tracking of user behavior on a webpage before making a purchase, or detailed sensor information collected about actions and environment while a person is driving a car. This abundance of information creates the perfect environment to leverage machine learning to its full potential. Indeed, powerful methods in machine learning and artificial intelligence such as gradient boosting and deep learning can achieve very strong predictive performance in a variety of tasks. However, these models are black boxes: it is near-impossible for a human to understand exactly how the input features are used to construct the predictions. In this regime of high-dimensional datasets, this limitation is particularly daunting, as it complicates understanding the relative quality of the various features that are being collected. In fact, it is often the case that most or all of the predictive performance can be achieved with a small subset of the features. This results in unnecessary computational complexity, more time spent training models and also lower performance out-of-sample, as an algorithm might not be able to correctly detect the features driving the signal when exposed to many noisy features. Unfortunately, we usually cannot directly determine such a subset of features a priori with intuition alone, and thus there has been great attention in the literature to develop data-driven methods that identify relevant features, permitting us to discard those that do not provide additional value to the prediction.

One of the most commonly used approaches for feature selection is based on assessing the variable importance of a machine learning model, which attempts to quantify the relative importance of each feature for predicting the target variable. The variable importance is calculated by measuring the incremental improvement in performance attributed to each use of a feature inside the model, and summarizing this information across the entire model. We can use this to identify those features that are deemed to have little or no importance and remove them from the model. Variable importance is also an approach for model explainability, whereby we attempt to understand how the model makes predictions by assessing which features are deemed important during model training.

However, it should come as no surprise that any method for feature selection can be useful only when it is also accurate. If a variable importance method falsely overestimates the importance of irrelevant features, it can lead to false discoveries and prevent elimination of spurious features. On the other hand, if the importance of relevant features is underestimated, we might discard important variables and negatively impact the performance of the final model. It is therefore critical to develop an understanding for how variable importance behaves, and in particular whether the variable importance exhibits different characteristics when applied to different machine learning models.

Tree-based models are some of the most commonly used approaches in machine learning, for both their power and interpretability. Single-tree models such as CART~\citep{breiman1984classification} are fully interpretable, as their prediction logic can be easily followed by observing the splits in the final decision tree. However, CART is trained using a greedy heuristic that forms the tree one split at a time, which has a number of downsides. First and foremost, this can result in trees that are far from globally optimal, as the best split at any given point in the greedy heuristic may not prove to be the best when viewed in the context of the future growth of the tree. Another key problem often cited in the literature is that the split selection method of CART is biased towards selecting features with a greater number of possible split points, due to the exhaustive search over all features at each step of the algorithm \citep{kononenko1995biases,kim2001classification, strobl2007unbiased}. These limitations are cause for concern when interpreting the variable importance of CART, as the selected features may be biased towards those with a greater number of unique values, and the greedy algorithm may lead to incorrect features being used in the splits near the root of the tree, which are usually those that receive most importance.

Tree-based ensemble methods such as random forests~\citep{breiman2001random} and gradient boosting (e.g. XGBoost~\citep{chen2016xgboost}) improve upon the performance of CART by combining the predictions of many such trees. This indeed leads to state-of-the-art performance, but sacrifices the interpretability of the model, as it is near-impossible to comprehend the behaviors and interactions of hundreds of decision tree models. As a result, it is common to rely on variable importance approaches to understand and explain these models. Similar to CART, it is known that the variable importance calculations of these models can be sensitive to the same issues of bias towards features with more potential split points. SHAP~\citep{NIPS2017_7062} is a recent method that unifies many earlier approaches aimed at resolving this bias issue, and uses a game-theoretic approach to understand and explain how each feature drives the final prediction. In recent years, SHAP has rapidly become widely-used for the purposes of explaining black-box models and conducting feature selection, in part due its perceived robustness and resolution of bias concerns.

As mentioned earlier, the predictive performance of CART suffers compared to ensemble methods, but the latter are forced to sacrifice the interpretability of a single decision tree in order to achieve this performance, forcing practitioners to choose between performance and interpretability. A recently stream of work seeks to address this issue by constructing decision trees with global optimization techniques rather than greedy heuristics. In particular, Optimal Trees~\citep{bertsimas2017optimal,bertsimas2019machine} utilizes mixed-integer optimization to construct decision trees in a single step that are globally optimal. The resulting model maintains the interpretability of a single decision tree, but has been shown to outperform CART and has performance competitive with black-box models. Since the method considers optimizing all splits in the tree simultaneously rather than one-by-one greedily, we might expect that the split selection is less susceptible to the same bias issues as CART.

In this paper, we investigate the performance of variable importance approaches for conducting feature selection across a variety of machine learning models, both black boxes like XGBoost (both alone and with SHAP) and those that are intrinsically interpretable like CART and Optimal Trees. To do so, we conduct a number of experiments with synthetically-generated data where the ground truth subset of relevant features is known, and measure the relative ability of different approaches to recover this truth. The results from these experiments provide compelling evidence for the following conclusions:
\begin{itemize}
\item XGBoost (both with and without SHAP) is ineffective as a method for feature selection. Even in very simple cases, significant importance is assigned to features that are in fact completely unrelated to the target.
\item Optimal Trees is the strongest performing method for feature selection. In all cases we consider, Optimal Trees is the least likely to assign importance to variable that are unrelated to the target, and is makes significantly more efficient use of data compared to CART.
\item While CART indeed suffers in the presence of biased data, the performance of Optimal Trees is unaffected, demonstrating that concerns of bias in split selection can be overcome through use of global optimization techniques.
\end{itemize}

The structure of the paper is as follows. In Section~\ref{experiment_1}, we conduct an experiment in which we generate data according to a fixed ground truth process with known variable importance, and measure the ability of methods to recover this truth. In Section~\ref{experiment_2}, we conduct an experiment in which we generate a number of randomized trees to serve as the ground truth, and compare the rate at which methods assign importance to irrelevant variables. In Section~\ref{experiment_3}, we repeat the experiment of Section~\ref{experiment_2}, but introduce significant variation in the number of unique values across features in order to assess the susceptibility of each method to biases in the data. Finally, we conclude in Section \ref{conclusion}.

\section{Experiment 1: Fixed regression tree}
\label{experiment_1}

In this section, we consider a scenario where the data is generated according to a fixed ground truth regression tree. Given data generated according to this tree structure, we can compare how well each method is able to recover the true variable importance.

\subsection{Setup}

We follow a similar setup to that used in \cite{hothorn2006unbiased}, where the authors propose a simple experiment to examine whether greedy tree methods are biased towards features with more unique values. We generate five uniformly distributed random variables $X_1$ , ... , $X_5$ $\sim U$[0, 1] serving as numeric covariates. To introduce variety in the number of unique values, we replace 25\% of the values in $X_4$ with missing values, and we round values of $X_5$ to one digit. We also add an additional binary feature, $X_6 \sim \text{Bernoulli}(0.5)$. As a result, all values of the first three features are unique, while $X_4$ has 75\% unique values, $X_5$ has 10 unique values, and $X_6$ has 2 unique values.

\begin{figure}
    \centering
    \includegraphics[width=\linewidth]{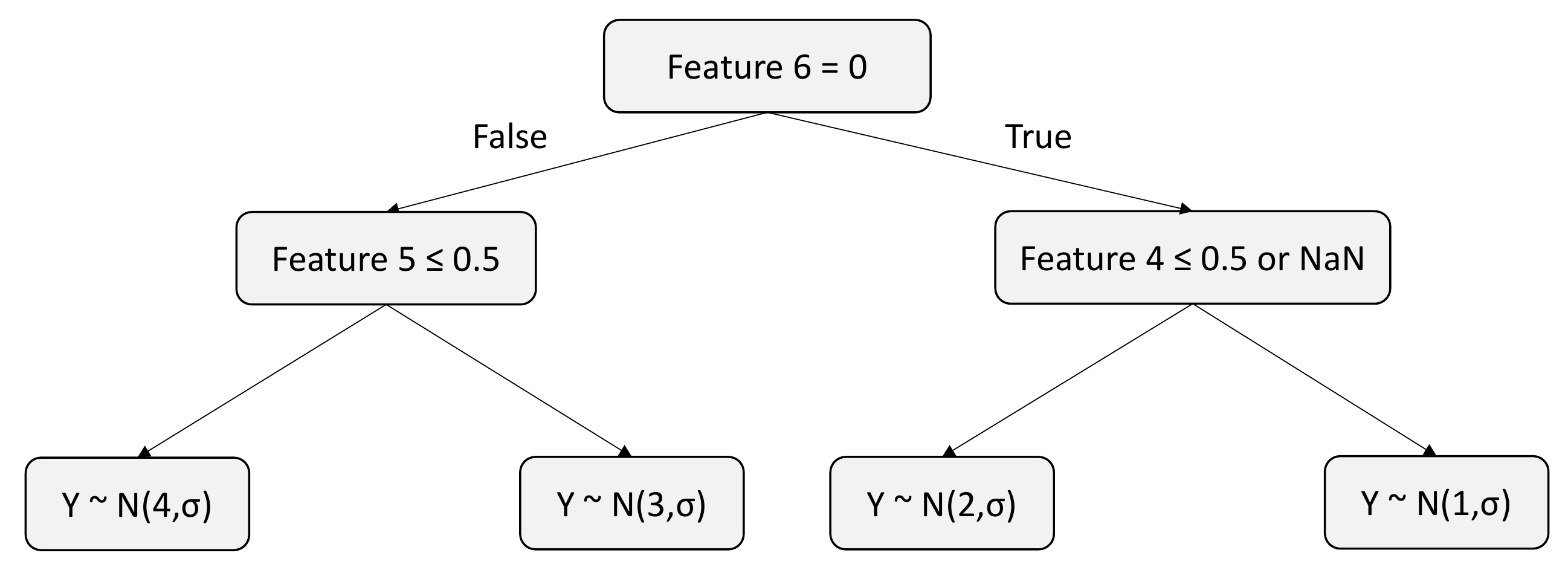}
    \caption{Ground truth tree for Experiment 1.}
    \label{fig:ground_truth}
\end{figure}

Next, we assume the dependent variable $Y$ takes the following form as a function of $X_4$, $X_5$ and $X_6$:
\begin{equation*}
  \textbf{Y} \sim
    \begin{cases}
      \textit{N}(1, \sigma) & \text{if $X_6$ = 0 and ($X_4$ $\leq$ 0.5 or $X_4$ is missing)}\\
      \textit{N}(2, \sigma) & \text{if $X_6$ = 0 and ($X_4$ $>$ 0.5 and $X_4$ not missing)}\\
      \textit{N}(3, \sigma) & \text{if $X_6$ = 1 and $X_5$ $\leq$ 0.5}\\
      \textit{N}(4, \sigma) & \text{if $X_6$ = 1 and $X_5$ $>$ 0.5}
    \end{cases}       
\end{equation*}
where $\sigma$ is a specified level of noise for the experiment. This structure is equivalent to a regression tree with three splits, as shown in Figure~\ref{fig:ground_truth}. To motivate the design of the experiment, the structure is such that the three features with more unique values are not related to the data generation process. If methods are indeed biased towards choosing these features, this will result in erroneously higher importance being assigned to these features.

We generate features and target data based on this ground-truth tree structure, for training set sizes from $n =$100--5000 points. We then run each method and calculate the variable importance of the resulting model. We use 25\% of the training data, and consider the following methods:

\begin{itemize}
    \item CART, validating over \texttt{cp} with cost-complexity pruning;
    \item Optimal Classification Trees (OCT), validating over \texttt{max depth} between 1 and 10 and \texttt{cp} with cost-complexity pruning;
    \item XGBoost, validating over \texttt{max depth} between 1 and 10 (we experimented with validating other parameters and found this had no impact on the results).
\end{itemize}

As an additional comparison, we also apply SHAP to the trained XGBoost model and extract the SHAP importance scores.

We consider two levels of noise ($\sigma = 1$ and $\sigma = 2$), and repeat the experiment 2000 times, reporting the average importance assigned to each feature by each method.

\subsection{Results}
\label{result_1}

Given that the ground truth tree is fixed, we know that $X_6$ should be the most important feature with importance of 0.8, followed by $X_4$ and $X_5$ with importance of 0.1 each, and all other features are unrelated to the target and should have zero importance. 

\begin{figure*}
    \centering
    \includegraphics[width=0.9\textwidth]{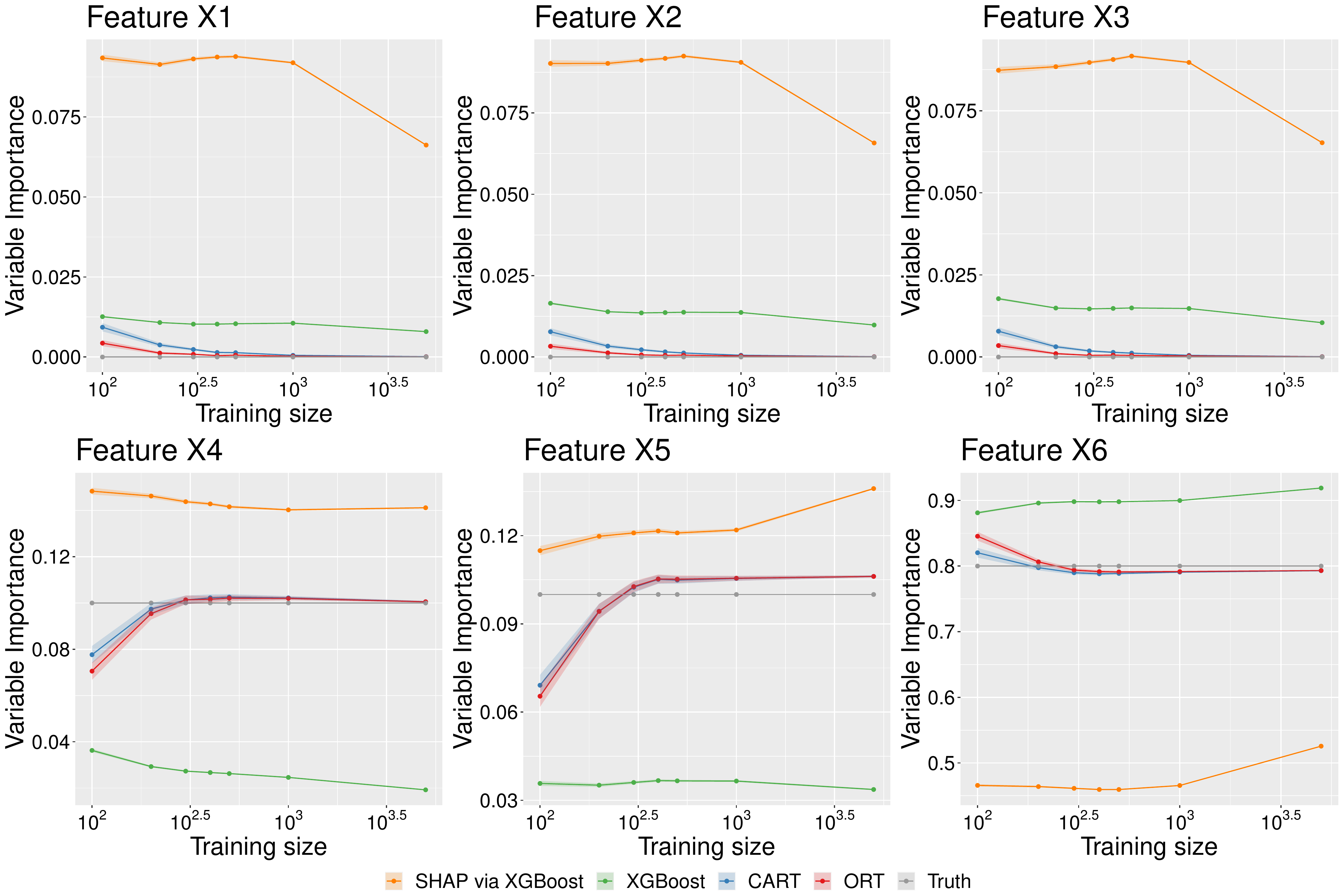}
    \caption{Average importance assigned to features in low noise setting ($\sigma = 1$).}
    \label{var_imp4_5}
\end{figure*}

First, we examine the behavior of each method as the number of data points increases. Figure~\ref{var_imp4_5} presents the results for the low noise setting ($\sigma = 1$). The first three features are quickly identified as irrelevant variables by both ORT and CART, with importance converging to zero faster for ORT than CART. In contrast, the importance for these features from XGBoost only decreases by small amounts as the number of points increases, and is still significantly above zero even for $n = 5000$. SHAP assigns even more importance to these features than XGBoost. For the remaining features, we see that ORT and CART again converge to the expected truth, whereas XGBoost and SHAP do not exhibit much change as the number of points is increased, and the values are significantly different to the expected true values.

\begin{figure*}
    \centering
    \includegraphics[width=0.9\textwidth]{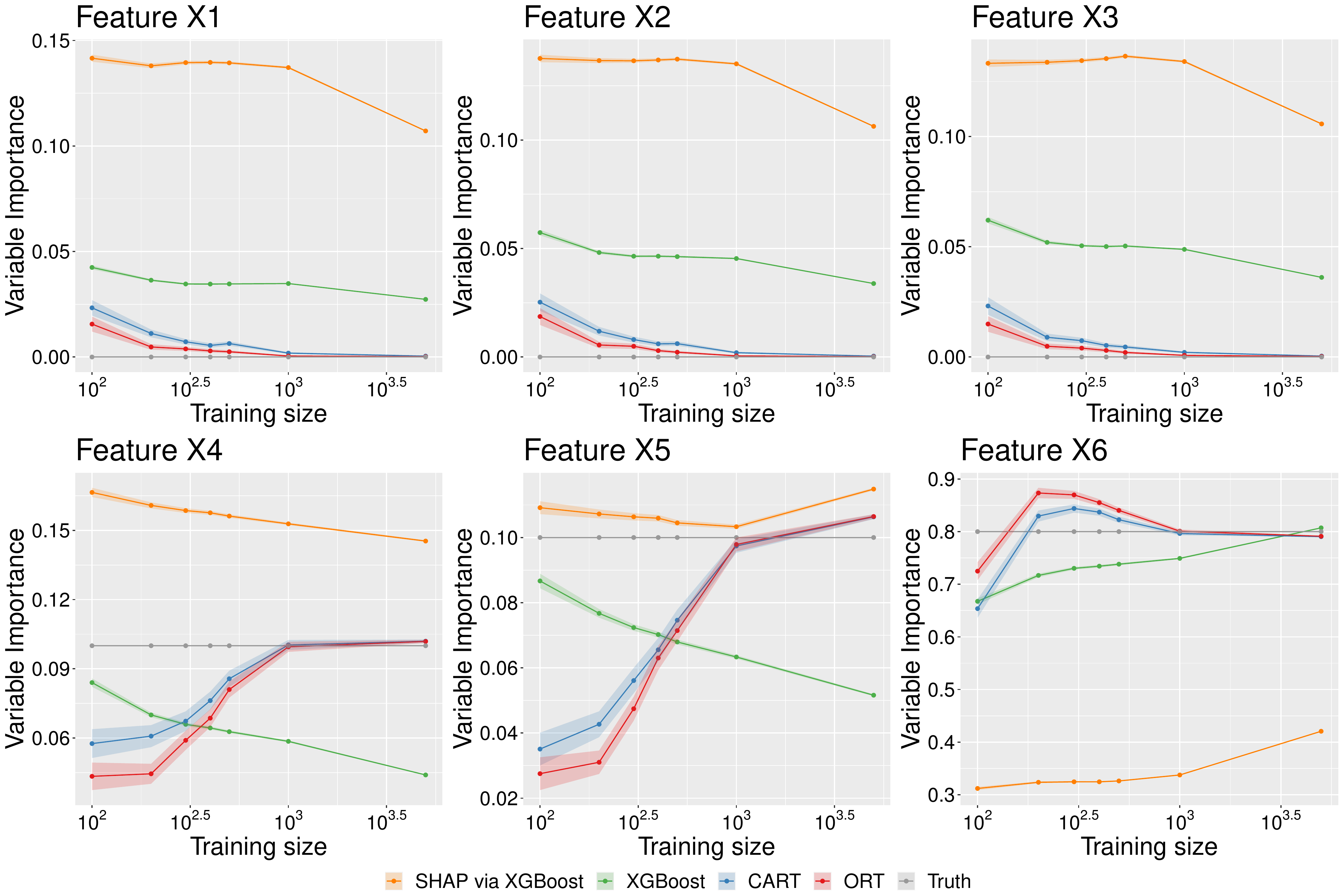}
    \caption{Average importance assigned to features in high noise setting ($\sigma = 2$).}
    \label{var_imp4_6}
\end{figure*}

Figure~\ref{var_imp4_6} presents the results in the high noise setting ($\sigma = 2$). Overall, we see similar trends to the low noise setting for CART and ORT, except that more data points are required for convergence to the expected values. ORT again exhibits faster convergence towards zero on the features that are unused by the model. XGBoost assigns similar importance to the first five features, placing most importance on the final feature, whereas SHAP underestimates the importance of the final feature and assigns greater importance to each of the first five features.

In order to assess the relative importance assigned to each feature, Tables~\ref{var_imp_std1} and \ref{var_imp_std2} summarize the variable importance results for the scenario with the most data ($n = 5000$). We can see that ORT and CART indeed converge to the expected values in both scenarios, but the same is not true for XGBoost and SHAP. Both XGBoost and SHAP consistently identify $X_6$ as the most important feature, but fail to identify that the first three features are unrelated to the target, as these importances are consistently significantly above zero. In the low noise setting, more importance is assigned to $X_4$ and $X_5$ than to the first three features, which is correct, but in the high noise setting, both XGBoost and SHAP return similar importance scores for the first five features. 
\begin{table}
\centering
\caption{Average importance assigned to features in low noise setting ($\sigma = 1$) with $n=5000$.}
\label{var_imp_std1}
\begin{tabular}{ccccc}
\toprule
Feature & ORT & CART & XGBoost & SHAP\\
\midrule 
$X_1$   &    0.0001   &   0.0001   &   0.0079   &   0.0662 \\
$X_2$   &    0.0001   &   0.0001   &   0.0098   &   0.0658 \\
$X_3$   &    0.0001   &   0.0001   &   0.0105   &   0.0653 \\
$X_4$   &    0.1005   &   0.1006   &   0.0192   &   0.1412 \\
$X_5$   &    0.1061   &   0.1062   &   0.0337   &   0.1360 \\ 
$X_6$   &    0.7932   &   0.7930   &   0.9189   &   0.5256 \\
\bottomrule
\end{tabular}
\end{table}

\begin{table}
\centering
\caption{Average importance assigned to features in low noise setting ($\sigma = 2$) with $n=5000$.}
\label{var_imp_std2}
\begin{tabular}{ccccc}
\toprule
Feature & ORT & CART & XGBoost & SHAP\\
\midrule 
$X_1$   &    0.0002   &   0.0004   &   0.0273   &   0.1071 \\
$X_2$   &    0.0001   &   0.0004   &   0.0339   &   0.1063 \\
$X_3$   &    0.0002   &   0.0004   &   0.0361   &   0.1057 \\
$X_4$   &    0.1019   &   0.1020   &   0.0440   &   0.1454 \\
$X_5$   &    0.1064   &   0.1063   &   0.0516   &   0.1149 \\ 
$X_6$   &    0.7912   &   0.7904   &   0.8072   &   0.4206 \\
\bottomrule
\end{tabular}
\end{table}

Based on the results of this simple experiment, there is a stark difference in the ability of different methods to identify which features are relevant for prediction. Both ORT and CART demonstrate convergence to the expected result in both low and high noise settings as the number of data points increases, with ORT being more efficient at identifying features that are not important. There is little evidence that the results of either methods are distorted by a bias towards features with more unique values, as they both manage to successfully identify that the features with more unique values are unimportant. In constrast, while XGBoost and SHAP successfully identify $X_6$ as the most important variable, they fail to recognize that the first three features are unrelated to the target, and in the high noise setting struggle to distinguish between the first five features entirely. 

\section{Experiment 2: Randomized classification trees}
\label{experiment_2}

The previous experiment considered a regression problem with a simple, fixed ground truth tree structure. In this section, we conduct experiments with randomly-generated ground truth trees, in order to examine the quality of variable importance as a feature selection method in a more complicated setting. 

\subsection{Setup}
\label{experiment_2_setup}

The setup for this experiment is adapted from experiments appearing in \cite{bertsimas2019machine} and \cite{murthy1995decision}. We generate ground truth classification trees with up to 15 splits by choosing split features and thresholds at random. The leaves of the resulting tree are then assigned one of two labels such that no two leaves sharing a parent have the same label, as otherwise this parent node could simply be replaced with a leaf of the same label without affecting the predictions of the overall tree.

To generate the training data, we first generate features uniformly at random, and then generate the target labels in accordance with the ground-truth tree structure. 

In order to pose this as a feature selection task, we generate data with $p=7$ features, but construct the ground truth tree using only three of these features. We therefore expect that the features that are not used in the ground truth tree should be assigned zero importance.

Similar to the previous experiment, we run each method and calculate the variable importance. We report the percentage of importance that is assigned to the features that were not used in the ground truth tree. As before, we conduct this experiment for training set sizes $n=$100--5000, and reserve 25\% of the training set for validating parameters in the same way as before, with the following methods:

\begin{itemize}
    \item CART, validating \texttt{cp} with cost-complexity pruning;
    \item Optimal Classification Trees (OCT), validating over \texttt{max depth} between 1 and 10 and \texttt{cp} with cost-complexity pruning;
    \item XGBoost, validating over \texttt{max depth} between 1 and 10 (we experimented with validating other parameters and found this had no impact on the results).
\end{itemize}
 
We repeat the experiment 100 times, reporting both the average percentage of importance assigned to the irrelevant features by each method, and the out-of-sample accuracy on a hold-out test set of $50000$ points generated in the same manner as the training set.

\subsection{Results}
\label{result_2_1}

\begin{figure}
    \centering
    \includegraphics[width=0.47\textwidth]{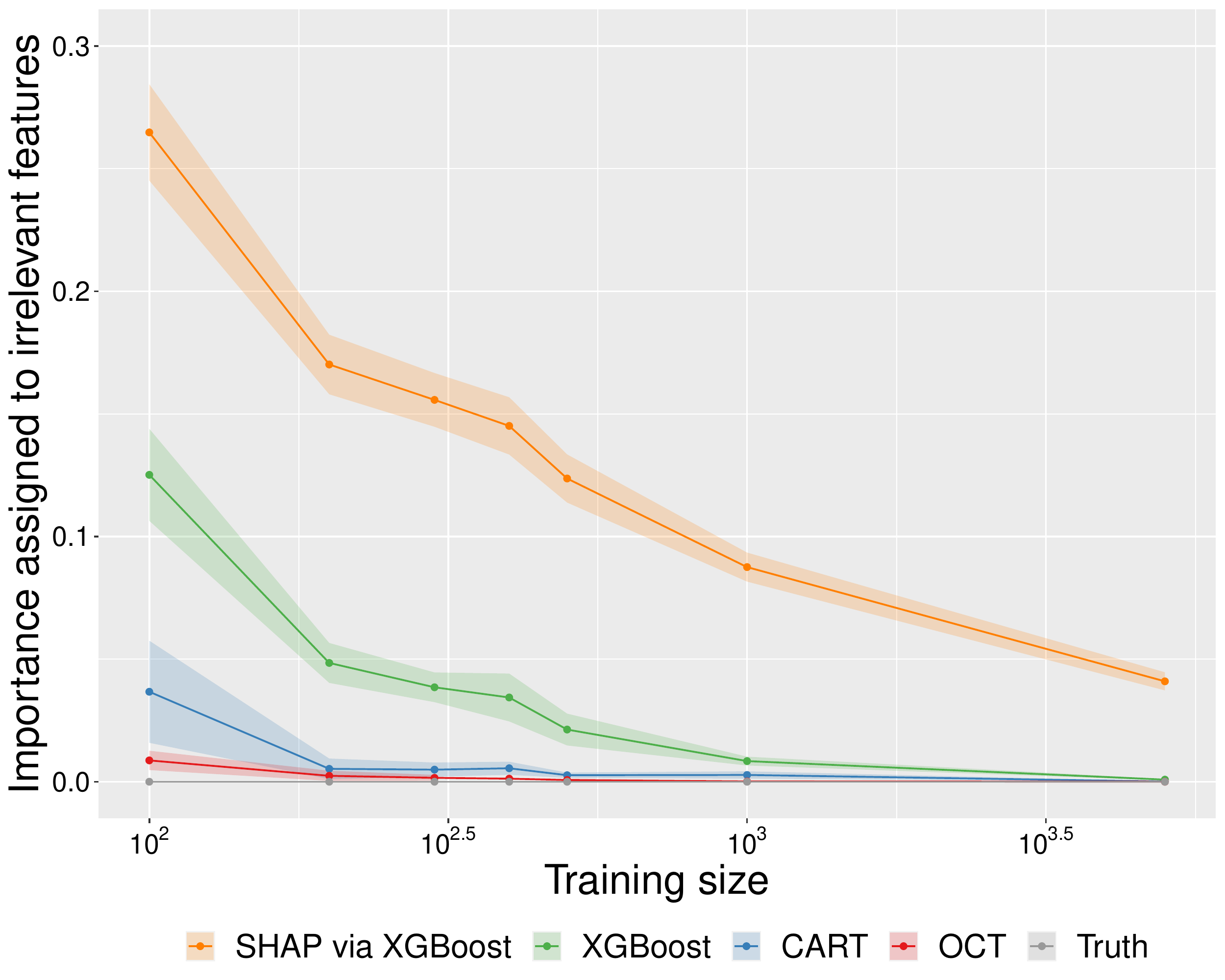}
    \caption{Model comparison in the no-bias setting for variable importance.}
    \label{fig:no_bias}
\end{figure}

Figure~\ref{fig:no_bias} shows the average importance assigned to the irrelevant features as the number of training points is increased. We see that OCT consistently assigns the least importance to these irrelevant variables among all the methods. CART performs weaker than OCT at small data sizes, but eventually reaches similar when the size of the training set is large. XGBoost and SHAP both assign significantly more importance to the unused features than CART and OCT. XGBoost eventually converges towards zero at $n=5000$, while SHAP needs significantly more data to do so.

The out-of-sample accuracy for each method is shown in Table~\ref{accuracy_unbiased}. We see that OCT and XGBoost perform similarly, both improving upon the performance of CART. This demonstrates that the improvements in importance detection offered by OCT did not come at the cost of any performance.
\begin{table}
\centering
\caption{Out-of-sample accuracy comparison in the no-bias setting.}
\label{accuracy_unbiased}
\begin{tabular}{ccccc}
\toprule
Sample Size & ORT & CART & XGBoost\\
\midrule 
100  & 0.9016   &   0.8641   &   0.8873 \\
 200  & 0.9529   &   0.9377   &   0.9515 \\
 300  & 0.9618   &   0.9490   &   0.9627 \\
 400  & 0.9736   &   0.9562   &   0.9706 \\
 500  & 0.9826   &   0.9675   &   0.9795 \\
1000 &  0.9916   &   0.9805   &   0.9893 \\
5000 &  0.9984   &   0.9966   &   0.9981 \\
\bottomrule
\end{tabular}
\end{table}

Altogether, this experiment reinforces the results of the previous experiment that XGBoost and SHAP struggle to correctly identify features as irrelevant compared to CART and Optimal Trees, and that the greedy method of CART is significantly less efficient at finding the correct answer compared to the global optimization used by Optimal Trees.

\section{Experiment 3: Randomized classification trees and biased data}
\label{experiment_3}

In this section, we repeat the experiment of Section~\ref{experiment_2} but modify the data generation process to include features with different numbers of unique values so that we can directly measure the degree to which each method might become biased towards features with more unique values, and what impact this has on the ability to select features accurately.

\subsection{Setup}

The setup of this experiment is identical to that described in Section~\ref{experiment_2_setup} with the exception of the mechanism for generating the features. Previously these features were generated uniformly at random, but now we introduce the potential for selection bias by rounding the last four features such that these features have 2, 4, 10, and 20 unique values, respectively. As before, each ground truth tree randomly selects three of these features that it uses for splitting. If methods are indeed susceptible to a selection bias based on the number of unique values, we expect to see an increased rate of importance being assigned to the first three features, regardless of whether they are in the tree.

\subsection{Results}
\label{result_2_2}

\begin{figure}[tb]
    \centering
    \includegraphics[width=0.47\textwidth]{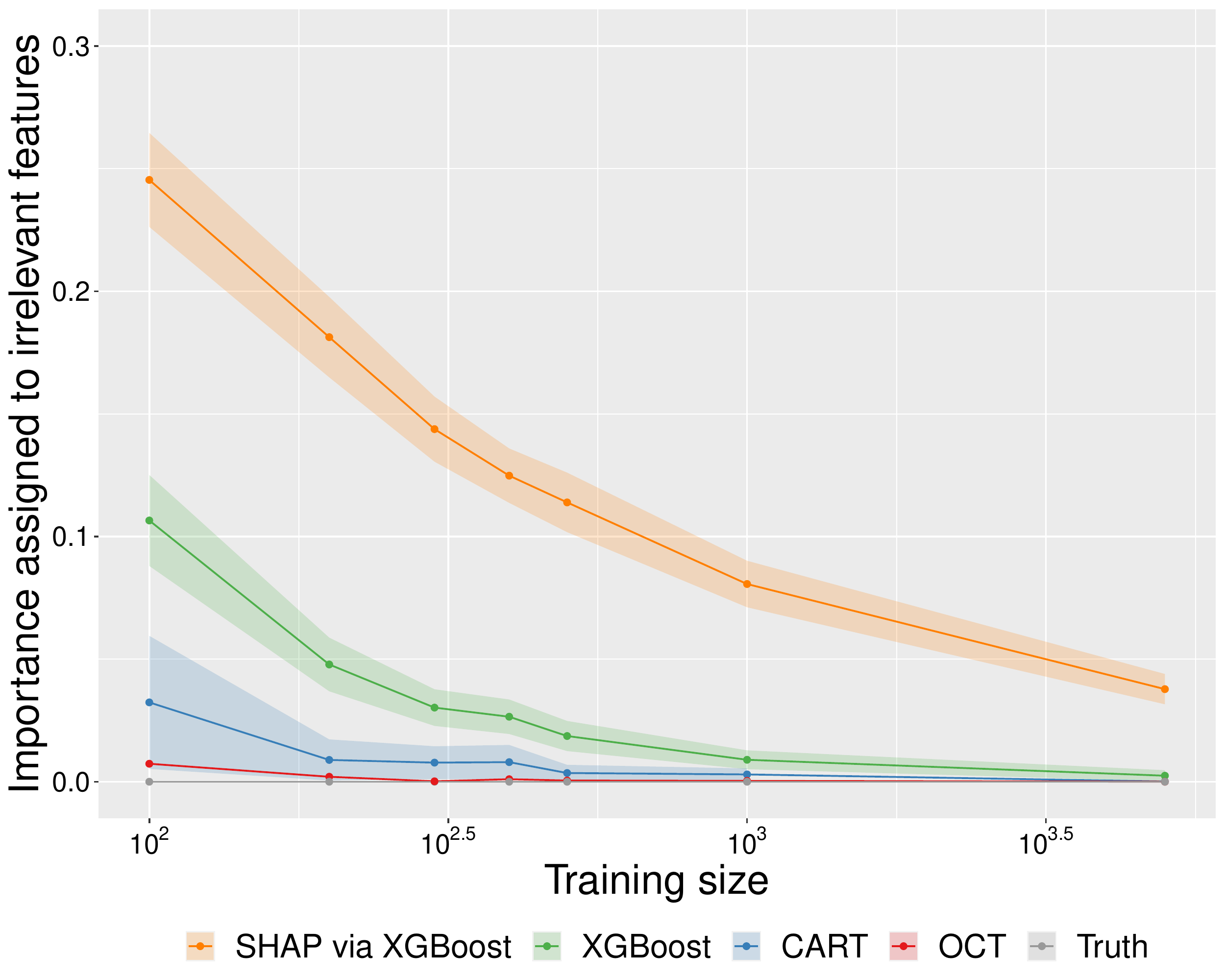}
    \caption{Model comparison in the biased setting for variable importance.}
    \label{fig:bias}
\end{figure}

Figure~\ref{fig:bias} shows the proportion of importance assigned to features unrelated to the data generation as the size of the training set increases. We see that again OCT is the strongest performing model and that the importance assigned to irrelevant features exhibits less variation compared to the previous experiment. On the other hand, while CART is still the second-best performer, we see that there is more variation in its results and it exhibits slower convergence than before. XGBoost and SHAP are still the weakest methods, performing roughly similar to before. 

The out-of-sample accuracy for each of the method is shown in Table~\ref{accuracy_biased}. We see a similar trend as before, with OCT and XGBoost performing similarly and outperforming CART, which again shows that OCT is not sacrificing any performance in exchange for the better identification of irrelevant features.

\begin{table}
\centering
\caption{Out-of-sample accuracy comparison in the biased setting.}
\label{accuracy_biased}
\begin{tabular}{ccccc}
\toprule
Sample Size & ORT & CART & XGBoost\\
\midrule 
100   &   0.9394    &   0.9120   &   0.9296\\
200   &   0.9780    &   0.9671   &   0.9731\\
300   &   0.9809    &   0.9664   &   0.9812\\
400   &   0.9888    &   0.9738   &   0.9853\\
500   &   0.9912    &   0.9789   &   0.9896\\
1000   &   0.9964    &   0.9868   &   0.9945\\
5000   &   0.9993    &   0.9944   &   0.9990\\
\bottomrule
\end{tabular}
\end{table}

This experiment shows that CART is indeed susceptible to a selection bias when faced with features that have varying numbers of unique values, as it exhibits more variability and slower convergence to the correct importance compared to before. However, even in this scenario, it performs better than XGBoost at selecting features. OCT appears to be unaffected by this bias and converges to the correct importance fastest among all methods, even at very small sample sizes, while still achieving the same predictive performance as XGBoost. SHAP is also unaffected by any selection bias, but is again unable to correctly identify that these features are not relevant for prediction.

\section{Conclusions}
\label{conclusion}

In this work, we investigated the relative performance of variable importance as a feature selection tool for a variety of machine learning algorithms. Currently, SHAP is widely used as an explainability method to make sense of the outputs of black-box models like XGBoost, and also for feature selection. On the other hand, interpretable methods such as decision trees are often considered flawed by design in the literature because of a bias towards features with more unique values.

Our simple experiments provide concrete evidence that challenges these beliefs. Our results show that SHAP and XGBoost consistently underestimate the importance of key features and assign significant importance to irrelevant features, and in the higher noise setting fail to distinguish between the two entirely. This raises serious concerns if these methods are used for feature selection or explanation, as they could lead us to drop a number of features that are useful, or alternatively to keep irrelevant features in the model and assign them significant importance when explaining the model. 

On the other hand, interpretable single-tree methods are shown to be very efficient in identifying which features are not relevant for prediction, driving their importance to zero with relatively little training data required. Additionally, we find that Optimal Trees exhibits faster identification of irrelevant features and less susceptibility to selection bias compared to CART as a result of its focus on global optimization.

This work provides evidence that despite widespread use, the variable importance scores from both XGBoost and SHAP may not be particularly effective at correctly determining feature importance. The results suggest exercising caution when using these approaches to understand the inner workings of black-box models, as it is impossible to tell if the features being assigned importance are actually relevant to the task at hand. In contrast, interpretable single-tree methods are fully transparent and effective at eliminating irrelevant features, and in the case of Optimal Trees, this often comes at little-to-no performance cost.

\bibliography{refs}

\begin{thebibliography}{11}
\providecommand{\natexlab}[1]{#1}
\providecommand{\url}[1]{\texttt{#1}}
\expandafter\ifx\csname urlstyle\endcsname\relax
  \providecommand{\doi}[1]{doi: #1}\else
  \providecommand{\doi}{doi: \begingroup \urlstyle{rm}\Url}\fi

\bibitem[Bertsimas and Dunn(2017)]{bertsimas2017optimal}
Dimitris Bertsimas and Jack Dunn.
\newblock Optimal classification trees.
\newblock \emph{Machine Learning}, 106\penalty0 (7):\penalty0 1039--1082, 2017.

\bibitem[Bertsimas and Dunn(2019)]{bertsimas2019machine}
Dimitris Bertsimas and Jack Dunn.
\newblock \emph{Machine learning under a modern optimization lens}.
\newblock Dynamic Ideas LLC, 2019.

\bibitem[Breiman(2001)]{breiman2001random}
Leo Breiman.
\newblock Random forests.
\newblock \emph{Machine learning}, 45\penalty0 (1):\penalty0 5--32, 2001.

\bibitem[Breiman et~al.(1984)Breiman, Friedman, Stone, and
  Olshen]{breiman1984classification}
Leo Breiman, Jerome Friedman, Charles~J Stone, and Richard~A Olshen.
\newblock \emph{Classification and regression trees}.
\newblock CRC press, 1984.

\bibitem[Chen and Guestrin(2016)]{chen2016xgboost}
Tianqi Chen and Carlos Guestrin.
\newblock Xgboost: A scalable tree boosting system.
\newblock In \emph{Proceedings of the 22nd acm sigkdd international conference
  on knowledge discovery and data mining}, pages 785--794, 2016.

\bibitem[Hothorn et~al.(2006)Hothorn, Hornik, and Zeileis]{hothorn2006unbiased}
Torsten Hothorn, Kurt Hornik, and Achim Zeileis.
\newblock Unbiased recursive partitioning: A conditional inference framework.
\newblock \emph{Journal of Computational and Graphical statistics}, 15\penalty0
  (3):\penalty0 651--674, 2006.

\bibitem[Kim and Loh(2001)]{kim2001classification}
Hyunjoong Kim and Wei-Yin Loh.
\newblock Classification trees with unbiased multiway splits.
\newblock \emph{Journal of the American Statistical Association}, 96\penalty0
  (454):\penalty0 589--604, 2001.

\bibitem[Kononenko(1995)]{kononenko1995biases}
Igor Kononenko.
\newblock On biases in estimating multi-valued attributes.
\newblock In \emph{Ijcai}, volume~95, pages 1034--1040. Citeseer, 1995.

\bibitem[Lundberg and Lee(2017)]{NIPS2017_7062}
Scott~M Lundberg and Su-In Lee.
\newblock A unified approach to interpreting model predictions.
\newblock In I.~Guyon, U.~V. Luxburg, S.~Bengio, H.~Wallach, R.~Fergus,
  S.~Vishwanathan, and R.~Garnett, editors, \emph{Advances in Neural
  Information Processing Systems 30}, pages 4765--4774. Curran Associates,
  Inc., 2017.
\newblock URL
  \url{http://papers.nips.cc/paper/7062-a-unified-approach-to-interpreting-model-predictions.pdf}.

\bibitem[Murthy and Salzberg(1995)]{murthy1995decision}
Sreerama~K Murthy and Steven Salzberg.
\newblock Decision tree induction: How effective is the greedy heuristic?
\newblock In \emph{KDD}, pages 222--227, 1995.

\bibitem[Strobl et~al.(2007)Strobl, Boulesteix, and
  Augustin]{strobl2007unbiased}
Carolin Strobl, Anne-Laure Boulesteix, and Thomas Augustin.
\newblock Unbiased split selection for classification trees based on the gini
  index.
\newblock \emph{Computational Statistics \& Data Analysis}, 52\penalty0
  (1):\penalty0 483--501, 2007.

\end{thebibliography}
\end{document}